\documentclass[11pt,a4paper]{article}
\usepackage[table,xcdraw]{xcolor}
\usepackage[hyperref]{emnlp2020}
\usepackage{times}
\usepackage{latexsym}
\usepackage{multirow}
\usepackage{mathrsfs}
\usepackage{amsmath}
\usepackage{amssymb}
\usepackage{graphicx}
\usepackage[normalem]{ulem}
\useunder{\uline}{\ul}{}
\usepackage{subcaption}
\usepackage{color}

\usepackage{enumitem}
\usepackage{microtype}

\aclfinalcopy % Uncomment this line for the final submission
\title{Plug-and-Play Conversational Models}

\author{Andrea Madotto$^1$\thanks{\hspace{1mm}Equal Contribution}, Etsuko Ishii$^{*1}$, Zhaojiang Lin$^{*1}$, Sumanth Dathathri$^{*}$\thanks{\hspace{1mm}Work done primarily at the Caltech.}, Pascale Fung$^{1}$ \\
  $^1$The Hong Kong University of Science and Technology\\
  \texttt{\{amadotto,eishii,zlinao,pascale\}@ust.hk} \\
   \texttt{sdathath@gmail.com}
  }

\date{}

\begin{document}
\maketitle
\begin{abstract}
There has been considerable progress made towards conversational models that generate coherent 
and fluent responses; however, this 
often involves training large language models on large dialogue datasets, such as Reddit.
These large conversational models provide little control over the generated 
responses, and this control is further limited in the absence of 
annotated conversational datasets for attribute specific generation 
that can be used for fine-tuning the model.
In this paper, we first propose and evaluate plug-and-play methods 
for controllable response generation, which does not require 
dialogue specific datasets and does not rely on fine-tuning a large model.
While effective, the decoding procedure induces considerable computational overhead,
rendering the conversational model unsuitable for interactive usage.
To overcome this, we introduce an approach that does not require 
further computation at decoding time, while 
also does not require any fine-tuning of a large language model. 
We demonstrate, through extensive automatic and human evaluation, a high degree of
control over the generated conversational responses with regard to multiple 
desired attributes, while being fluent.\footnote{Code available at \url{https://github.com/andreamad8/PPCM}}

\end{abstract}

\begin{table}[t]
    \centering
    \resizebox{\linewidth}{!}{
        \begin{tabular}{rl}
        \hline
        \textit{\textbf{\small HUMAN 1}} &
          \begin{tabular}[c]{@{}l@{}}Hmmmm. I really want to try bull riding. \\ Do you have any interest in that?\end{tabular} \\
        \textit{\textbf{\small HUMAN 2}} &
          \begin{tabular}[c]{@{}l@{}}I'd love to try! Can we schedule something\\ for next week?\end{tabular} \\ \hline
        \cellcolor[HTML]{CECECE} 
        \textit{\textbf{\small DGPT}} &
          \cellcolor[HTML]{F2F2F2}Yes! Let's get some stuff going. \\
        \cellcolor[HTML]{8F98C9} 
        \textit{\textbf{\small NEGATIVE}} &
          \cellcolor[HTML]{D9E3FF}\begin{tabular}[c]{@{}l@{}}I'm actually going to try my first ever bull \\ riding weekend in July, in the middle of \\ a mud sludge of mudslide.\end{tabular} \\
        \cellcolor[HTML]{F6A8A8} 
        \textit{\textbf{\small POSITIVE}} &
          \cellcolor[HTML]{FFF0F7}\begin{tabular}[c]{@{}l@{}}That would be wonderful! I'm excited to\\ see you guys again in a couple of weeks :D\end{tabular} \\
        \cellcolor[HTML]{FDDEAF} 
        \textit{\textbf{\small QUESTION}} &
          \cellcolor[HTML]{FDEFD9}\begin{tabular}[c]{@{}l@{}}What's it like? Do you ride on a regular\\ basis or just for fun?\end{tabular} \\
        \cellcolor[HTML]{C5DCB6} 
        \textit{\textbf{\small BUSINESS}} &
          \cellcolor[HTML]{E1ECD9}\begin{tabular}[c]{@{}l@{}}Yea sure i can wait until after work. I am off \\ at the end of my shift in the middle east.\end{tabular} \\
        \cellcolor[HTML]{F3F0E7} 
        \textit{\textbf{\begin{tabular}[c]{@{}l@{}}{\small SCIENCE}\\ {\small \& TECH}\end{tabular}}} &
          \cellcolor[HTML]{F7F7F3}\begin{tabular}[c]{@{}l@{}}Sure thing! I'll PM you my email. We can \\ set up a dropbox or something and just \\ message people together if you'd like.\end{tabular} \\
        \cellcolor[HTML]{B8D8E1} 
        \textit{\textbf{\small SPORTS}} &
          \cellcolor[HTML]{E0EEF2}\begin{tabular}[c]{@{}l@{}}Ok how about Wednesday at 7pm central? \\ I'll be there with a buddy of mine \\ who is a bull rider.\end{tabular} \\
        \textit{\textbf{\small HUMAN 1}} &
          Sure! What does your Saturday look like? \\ \hline
        \end{tabular}
    }
    \caption{Example of controllable response generation. DGPT is the DialoGPT~\cite{zhang2019dialogpt} response, while the others are generated using different attribute models.}
\end{table}
\section{Introduction}
\label{sec:intro}
Large pre-trained language models~\cite{peters2018deep,radford2019language,raffel2019exploring} have greatly improved the state-of-the-art in many down-stream tasks.
These language models are trained using the simple log-likelihood objective over large amounts of unlabeled data (e.g., Wikipedia articles).
This approach results in large powerful language 
models that produce coherent text and can be used to perform unconditional language generation. 
However, they provide little control over the 
text generated. 
In the context of conversational models, \citet{see2019makes} show that being 
able to control the response generation can have a significant impact 
on the quality of conversations. 
However, controlled generation from these large conversational 
models remains a challenge, and is particularly more difficult 
in the absence of annotated conversational datasets.

%
% More challenging is to control the output of the generation in terms of topics and style, especially without using supervised datasets. 

For large language models, controlled generation has recently received increased attention. 
In CTRL~\cite{keskar2019ctrl}, the language model is trained 
to generate based on a control code presented to the model at 
the start of the context.
In~\citet{ziegler2019fine}, GPT-2 \citep{radford2019language} is fine-tuned using reinforcement-learning with human annotators in the 
loop to generate contuining text with positive sentiment.
Both of these approaches require learning/fine-tuning all of the models' parameters, and new desired attributes cannot be 
easily incorporated into the generation once the models have 
been trained. 
Other approaches that do not alter the language model, but 
modify the decoding procedure for controlled generation include 1) re-weighting the output distribution using discriminators~\cite{holtzman2018learning} or bag of words~\cite{ghazvininejad2017hafez,see2019makes,baheti2018generating}, and 2) perturbing the models activation with an attribute model (PPLM)~\cite{dathathri2019plug}. 
These approaches, instead, are plug-and-play methods in that they can be used on top of any existing pre-trained language model. These methods, do not modify or train the parameters of the original models and they can achieve comparable performance to fine-tuning methods~\cite{dathathri2019plug}. Weighted decoding is generally difficult to tune because it can easily generate unrelated responses when the weight is not properly set \cite{see2019makes}. On the other hand, \cite{dathathri2019plug} incurs a high computational cost during the decoding stage, which is problematic for online systems such as dialogue systems. 

Open-domain conversational systems are a special case of language models where the prefix is the dialogue history and the continuation is a human-like response~\cite{wolf2019transfertransfo}. Recently, large pre-training language models trained on unlabeled human-to-human conversation (i.e. Reddit)~\cite{zhang2019dialogpt,adiwardana2020towards,roller2020recipes} have shown excellent performance in modelling human responses. Similarly, the output of large pre-trained conversational models cannot be directly controlled without having to re-train/fine-tune the model from scratch, which is practically inconvenient and sometimes impossible since few or no-conversational datasets exist for certain attributes or styles. 

On the other hand, plug-and-play methods are a viable solution since they do not require dialogue specific datasets, and they can be computed online on top of existing pre-trained models. A major drawback however is the high computational cost~\cite{dathathri2019plug} at decoding time.
This is acceptable for language models, where generating paragraphs or stories can be done offline, but it is problematic for online systems such as conversational models.
In this paper, we explore the approach from \citet{dathathri2019plug} (PPLM) in large pre-trained dialogue models for controlling the style and topic of the responses without fine-tuning on any dialogue specific dataset. Moreover, to cope with the computational cost at the decoding time, we propose to generate style/topic consistent responses with PPLM~\cite{dathathri2019plug} and then use it to optimize residual adapters~\cite{houlsby2019parameter} for directly learning how to steer the original distribution towards the selected attribute. 

With our extensive automatic and human evaluation, we empirically demonstrate that plug-and-play methods are effective in controlling the response while being computationally efficient. To summarize, our key contributions are:
% \{Maybe add a line about how 
% this isn't just fine-tuning an adapter, and 
% say we use synthesis data or something? 
% Maybe something like... to overcome 
% the comp cost, we show first generating a small 
% synthetic data and then fine-tuning is a
% viable alternative. The point here is...
% pitch generation + adapter-finetuning as 
% solution. Don't undersell saying we just fine-tuned.}
%
% To summarize, our key contributions are:
\begin{itemize}[leftmargin=*]
    \item we show the effectiveness of plug-and-play methods in large pre-trained conversational models using a variety of styles and topics such as Positive, Negative, Question, Sport, Business/Finance, without using dialogue specific dataset.
    % \item to overcome the computational cost of plug-and-play methods during inference, we show that residual adapters~\cite{houlsby2019parameter} are a viable solution for real time systems;
    % \item to overcome the computational cost during inference, we first generated a small set of synthetic responses from PPLM, and then we optimize the parameters of a residual adapter~\cite{houlsby2019parameter} to make the controllable response generation viable for online systems.
    \item we propose to use residual adapters~\cite{houlsby2019parameter}, which adds less than 1.5\% task-specific parameters per style/topic, to make the controllable response generation viable for online systems. 
    % to overcome the computational cost during inference, we propose to generate attribute consistent responses with PPLM~\cite{dathathri2019plug} and then use it to optimize tiny residual adapters~\cite{houlsby2019parameter} for steering the original distribution.
    % we first generated a small set of synthetic responses from PPLM, and then we optimize the parameters of a residual adapter~\cite{houlsby2019parameter} to make the controllable response generation viable for online systems.
    % \item we propose to use residual adapters~\cite{houlsby2019parameter} to make the controllable response generation viable for online systems.
    % \{Same comment as before: 
    % Solution = synth data + fine-tine}
    %, which adds less than 1\% task-specific parameters per style/topic
    \item we run a comprehensive automatic and human evaluation to show that plug-and-play methods can control the generate responses in term of style and topics, without losing fluency.  
    % \{What is attribute coherent response? You mean "coherent/fluent responses while being consistent with attribute. What is both 
    % coherent and fluent here? Grammar is wrong. Please restructure."} responses while being highly fluent;
    \item we carry out a thorough qualitative analysis on the difficulty of steering conversational models, highlighting current limitations and possible solutions.
    % \{I don't like 'unvieling' here. 
    % Can you restructure, 'unveiling current limitations and possible solutions'... 
    % I don't have better alternatives. Will 
    % add if I can think of something nicer to 
    % say}
    
\end{itemize}

\section{Related work}
\textbf{Open-domain conversational models} \ Generating human-like responses involves overcoming a variety of challenges such as personalization~\cite{li2016persona,personachat,dinan2019second,wolf2019transfertransfo,madotto2019personalizing}, knowledge grounding~\cite{dinan2018wizard,gopalakrishnan2019topical,ghazvininejad2018knowledge,moghe2018towards,wu2020controllable}, emotions~\cite{li2017dailydialog,rashkin2018know,zhou2018emotional,fan2020facial,li2020empathetic}, diversity~\cite{li2016diversity,li2016deep,ghandeharioun2019approximating,serban2017hierarchical,gao2018neural} and so on. In terms of controlled dialogue generation, \citet{see2019makes} studied of conditional generative models~\cite{kikuchi2016controlling} and weighted decoding~\cite{ghazvininejad2017hafez} in controlling models trained on persona-chat. \citet{see2019makes} concluded that controlling specificity, relatedness, and repetition increase human-engagement, motivating us to extend the controllabitly to styles and topics. In this paper, we focus on these two since large pre-trained models can already achieve a high humanness score~\cite{adiwardana2020towards,roller2020recipes,zhang2019dialogpt}. 
% Models trained on large human-to-human datasets can achieve a high humanness score~\cite{adiwardana2020towards,roller2020recipes,zhang2019dialogpt} without tricks. 
%
% \{Remove all "Indeed,"... I haven't seen 
% this elsewhere...}
% \{See et al. also conclude how control 
% can increase human-engagement, maybe 
% use that to sell this work a bit here?}
% \{Maybe discuss how conditional generation 
% means you need to retrain, etc? Not sure if it's 
% done elsewhere already}

\quad \\
\textbf{Controlled Text Generation} Recent methods for controlled generation include fine-tuning models using supervised learning~\cite{peng2020few,subramani2019can}, reinforcement learning~\cite{ziegler2019fine}, adversarial training~\cite{yu2017seqgan}, by pre-training models with control codes~\cite{keskar2019ctrl,ficler2017controlling,chan2020cocon}, and other various approaches~\cite{zhang2020pointer,sheng2020towards,carbone2020etc}. Alternatively, weight decoding using both bag-of-words~\cite{holtzman2018learning,ghazvininejad2017hafez,baheti2018generating,see2019makes} and discriminators~\cite{holtzman2018learning,krause2020gedi}, does not require any fine-tuning. Similarly, \citet{dathathri2019plug} propose the Plug-and-Play Language Model (PPLM) to control the generation of a pre-trained language model, e.g., GPT2~\citep{radford2019language}, both in terms of style and topic of the generated text. Finally, residual adapters~\cite{houlsby2019parameter} has been used to learn multiple language generation tasks~\cite{lin2020exploring} without fine-tuning the original models' parameters.

Concurrently to our work, \citet{Smith2020ControllingSI} compare the performance and tradeoffs of three existing controllable language generation methods on 200 possible styles.

%
% This method uses the discriminator to update
% the model's activations during generation as in~\citet{nguyen2017plug}.

% \{There's an earlier paragraph discussing 
% WD and PPLM. Can we avoid duplication?} We could but we need to fulfill the space

\section{Methodology}
\label{sec:method} 
A dialogue consists of one or more alternating turns between two speakers. We define the dialogue history at turn $t$ as $\mathcal{D}_{t}=\{ U_1, S_1, \dots,U_{t}\}$ where $U_t$ is the user utterance and $S_t$ is the system response. For simplicity, we overload $\mathcal{D}_{t}$ to denote the concatenation of sequences across 
turns with a special token separating the turns. In this paper, we model the dialogue responses using a Transformer~\cite{vaswani2017attention}-based Language Model (\texttt{LM}) by using the dialogue history $\mathcal{D}_t$ as a prefix and then generating the continuation $S_t$ in an auto-regressive manner~\cite{DBLP:journals/corr/abs-1901-08149}. 
\paragraph{Causal Language Modeling} Let us denote the concatenation of $\mathcal{D}_t$  and $S_t$ as the sequence of tokens $X= \{x_{0},\dots,x_{n}\}$, then we can compute the language model distribution using the chain rule of probability~\cite{bengio2003neural} as:
\begin{equation}
    p(X) =  \prod_{i=1}^{n}p(x_i|x_{0}, \cdots, x_{i-1}).
\end{equation}
% \paragraph{Causal Language Modeling} Let us define the words in $S_t$ as $s_{1},\dots,s_{n}$, then we can factorize the language model distribution using the chain rule of probability~\cite{bengio2003neural} as:
% \begin{equation}
%     p(S_t|\mathcal{D}_t) =  \prod_{i}^{n}p(s_i|s_{<i},\mathcal{D}_t)
% \end{equation}
Following the notation of \citet{dathathri2019plug}, we define the transformer decoding process in a recursive manner. Let us define the matrix $H_t$ as the key-value pairs from the dialogue history past, i.e., $H_t = [(K_{t}^{(1)}, V_{t}^{(1)}),\cdots, (K_{t}^{(l)},
V_{t}^{(l)})]$, where $(K_{t}^{(i)}, V_{t}^{(i)})$ corresponds to the key-value pairs from the $i$-th layer generated at all time-steps from 0 to $t$.
Thus, we define the recurrent decoding process as:
\begin{equation}
    o_{t+1}, H_{t+1} = \texttt{LM} (x_{t}, H_t)
\end{equation}
and then $x_{t+1}$ is sampled from the distribution $p_{t+1}= \texttt{Softmax}(W o_{t+1})$, where $W$ is a linear transformation that maps the hidden state of the last layer $o_{t+1}$ to a vector of vocabulary size. This efficient transformer implementation~\citep{huggingface} leverages the cached memories to generate $x_{t+1}$ without recomputing $H_t$. 
% \Sumanth{Can 
% someone compare/check overlap with the PPLM paper? Don't want the paper to get into 
% trouble... if it's not, OK.. can easily paraphrase}.
\begin{table*}[t]
\centering
\begin{tabular}{r|c|c|cc|ccc}
\hline
\multicolumn{1}{c|}{\multirow{2}{*}{\textbf{Dataset}}} & \multirow{2}{*}{\textbf{Task}} & \multirow{2}{*}{\textbf{\#C}} & \multicolumn{2}{c|}{\textbf{Samples}} & \multicolumn{3}{c}{\textbf{F1-Score}} \\ \cline{4-8} 
\multicolumn{1}{c|}{} &  &  & \textit{Train} & \textit{Test} & \textit{Train} & \textit{Test} & \multicolumn{1}{l}{\textit{SotA}} \\ \hline
\textit{SST-5}~\cite{socher2013recursive} & Sentiment & 5 & 318,582 & 2210 & 77.68 & 47.01 & 55.50$\dagger$ \\ \hline
% \textit{Toxic Challenge}~\cite{toxicchallenge} & Toxic & 2 & 100,529 & 11,170 & 95.92 & 95.91 & 98.85$\ddag$ \\ \hline
\textit{Daily Dialogue}~\cite{li2017dailydialog} & Act & 4 & 92,650 & 10,295 & 80.58 & 80.00 & 86.10$\ddag$ \\ \hline
\textit{AG NEWS}~\cite{zhang2015character} & Topic & 4 & 120,000 & 7,600 & 90.68 & 90.65 & 95.44$\mathsection$ \\ \hline
\end{tabular}
\caption{Attribute dataset statistics and performance. State-of-the-Art (\textit{SotA}) results are taken from $\dagger$~\cite{munikar2019fine}, $\ddag$~\cite{kumar2019practical}, and $\mathsection$~\cite{yang2019xlnet}.}
\label{tab:discriminator}
\end{table*}
\subsection{Plug-and-Play Language Models}
\label{subsec:PPLM}
PPLM~\cite{dathathri2019plug} uses an attribute model (i.e., a classifier) for controlling the generated text. We denote the attribute model as $p(a|X)$ where $a$ is the specific desired attribute to optimize for (e.g., positivity), and $X$ is the generated response so far. At every generation step $t$, PPLM perturbs the history matrix $H_t$ in the direction of the sum of two gradients: i) to maximize the log-likelihood of the attribute $a$ under the conditional attribute model $p(a|X)$ and ii) ensuring high log-likelihood of the generated text under the unmodified conversational language model $p(X)$. 
The gradient updates are restricted to $H_t$ so to preserve the original model parameters.

Let $\Delta{H}_t$ be the update to $H_t$ to shift the generated text towards possesing the desired attribute $a$ i.e., $o_{t+1}, H_{t+1} = \texttt{LM} (x_{t}, H_t + \Delta{H}_t)$. At the beginning of the generation, $\Delta{H}_t$ is initialized to zero and it is updated using the gradients from the attribute model. Following \citet{dathathri2019plug}, we rewrite the attribute model $p(a|X)$ as $p(a|H_t + \Delta{H}_t)$ and we define the gradient update for $\Delta{H}_t$ as
\begin{equation}
      \Delta{H}_{t} \leftarrow \Delta{H}_{t} + \alpha \frac{\nabla_{\Delta{H}_{t}} \log p(a|H_t + \Delta{H}_t)}
        {\| \nabla_{\Delta{H}_{t}} \log p(a|H_t + \Delta{H}_t) \|^{\gamma} }
        \label{pplm}
\end{equation}
where $\alpha$ is the step size, and $\gamma$ is the scaling coefficient for the normalization term. Equation~\ref{pplm} is repeated $p$ times depending on how strongly we want the response to be conditioned to the attribute. We study the effect of the step-size $\alpha$ and the number of iterations $p$ on the generated text in detail in Section~\ref{sec:analysis}. 
Subsequently, the new $\widetilde{H}_t = H_t + \Delta{H}_t$ is computed and a new token is generated using $\widetilde{o}_{t+1}, H_{t+1} = \texttt{LM} (s_{t}, \widetilde{H}_t)$. The described optimization process is repeated for every token in the generated sequence. As aforementioned, to ensure fluency we also take a step towards minimizing the Kullback–Leibler (KL) regularization between the perturbed and the original distribution. In addition, we also use the Post-norm Geometric Fusion~\cite{stahlberg2018simple,dathathri2019plug} for avoiding adversarial generation~\cite{szegedy2013intriguing}.

% \Sumanth{This was briedly mentioned in the first paragraph in this section. So, maybe you want to connect ot that and say... as mentioned above, to ensure fluency, we 
% also take steps towards miniminzing the KL between the .... In addition, we also 
% use the Post-norm gemoetric fusion (cite PPLM and the cold-fusion paper). See PPLM for 
% a reference.}
% To ensure fluency and avoid adversarial generation~\cite{szegedy2013intriguing}, we followed \citet{dathathri2019plug}, which proposed to add a Kullback–Leibler (KL) regularization between the perturbed and the original distribution, and to merge the latter two using Post-norm Geometric Fusion~\cite{stahlberg2018simple}. 

\paragraph{Attribute Models}
In PPLM the authors propose two attribute models, such as bag-of-words and discriminators. In this paper, we focus on the latter, since discriminators based attribute models do not require human selected keywords.
% which may introduce bias\Sumanth{Most datasets 
% have biases in them as well, can you justify with something better.. I feel this is something
% Reviewer 2 will pick up easily}. 
%
The discriminator is a linear classifier $f$ trained on an annotated dataset with sentence and label pairs as $(x,y)$ -- note that these sentences do not necessarily need to be conversational responses, as in our case. 
For each sentence $x$ of length $t$, we compute the set of hidden states $o^x_{:t}$ from the \texttt{LM}, then we compute the mean ($\bar{o}^{t}$) across time, and finally we train $f$ using the cross-entropy between the label distribution $y$ and $f(\bar{o}^{t})$. 
% \Sumanth{cross entropy is between two distributions? 
% Change wording here maybe}.  

% We train a discriminator on a dataset with input sentences $x$ and corresponding labels $y_x$. For an input $x$ of length $t$, we compute $o^x_{:t}$ and train $f$ on the mean ($\bar{o}^{t}$) of the embedding across time.

\begin{table*}[t]
\centering
\begin{tabular}{rcccccccccc}
\multicolumn{1}{l}{} & \multicolumn{1}{l}{} & \multicolumn{1}{l}{} & \multicolumn{1}{l}{} & \multicolumn{1}{l}{} & \multicolumn{5}{c}{\textbf{Score by Attribute}} \\ \hline
\multicolumn{1}{c|}{\textbf{}} & $\downarrow$  \textbf{Ppl.}& $\uparrow$ \textbf{Dist 1/2/3} & \textbf{Discrim.} & \multicolumn{1}{c|}{\textbf{Score}} & \textbf{Posi.} & \textbf{Nega.} & \multicolumn{1}{l}{\textbf{Busin.}} & \multicolumn{1}{l}{\textbf{Sci/Tech}} & \multicolumn{1}{l}{\textbf{Sport}} \\ \hline
\multicolumn{1}{r|}{\textit{DG}} & \textbf{39.60} & 0.22/0.64/0.77 & 46.48 & \multicolumn{1}{c|}{32.91} & 65.67 & 19.40 & 17.41 & 91.04 & 27.86 \\
\multicolumn{1}{r|}{\textit{WD}} & 53.03 & 0.25/0.74/\textbf{0.84} & 50.18 & \multicolumn{1}{c|}{34.54} & 58.21 & 28.86  & 19.40 & 91.04 & 36.82 \\
\multicolumn{1}{r|}{\textit{PP}} & 45.86 & 0.24/0.67/0.79 & 73.28 & \multicolumn{1}{c|}{49.54} & 75.12 & 51.74  & 47.26 & 93.03 & 59.20 \\
\multicolumn{1}{r|}{\textit{AD}} & 41.57 & 0.17/0.58/0.77 & \textbf{96.52} & \multicolumn{1}{c|}{\textbf{70.01}} & \textbf{93.03} & \textbf{73.13} & \textbf{68.66} & \textbf{99.00} & \textbf{83.08} \\ \hline
% \multicolumn{1}{r|}{\textit{HM}} & 49.29 & \textbf{0.32}/\textbf{0.75}/0.83 & - & \multicolumn{1}{c|}{22.67} & 45.27 & 10.95 & 2.99 & 91.04 & 5.97 \\ \hline
\end{tabular}
\caption{Automatic evaluation results. In all the metrics higher is better except for Perplexity (Ppl.), and \textit{Discrim.} is the accuracy of the internal attribute model, while \textit{Score} is the accuracy of the external classifier. All the results, are averaged among the six attribute models.} \label{Tab:auto}
\end{table*}
\subsection{Residual Adapters}
\label{subsec:Adapters}
Residual Adapters~\cite{houlsby2019parameter,bapna2019simple} are trainable modules added on top of each transformer layer, which steer the output distribution of a pre-trained model without modifying the original weights. 
An adapter block consists of a Layer Normalization~\cite{ba2016layer} for efficient adaptation, followed by an auto-encoder~\cite{hinton1994autoencoders} with a residual connection. 
Formally, given the hidden representation at layer $i$ denoted as $o^i_{:t} \in \mathbb{R}^{t\times d}$, where $d$ is the hidden size and $t$ is the current generation step, the residual adapter computes:
\begin{align}
f_{\theta_i}(x) = \texttt{ReLU} \big ( &\texttt{LN}(x) \cdot W_i^{E} \big) \cdot W_i^{D}, \nonumber \\ 
\texttt{Adapter}(o^i_{:t}) &= f_{\theta_i}(o^i_{:t}) + o^i_{:t},
\end{align}
where $W_i^{E}$ and $W_i^{D}$ are trainable parameters of dimensions $d\times m$ and $m\times d$ respectively, and \texttt{LN}$(\cdot)$ denotes the layer normalization. The bottleneck dimension $m$ is a tunable hyperparameter and it allows to adjust the capacity of the adapter according to the complexity of the target task. 
We denote $\theta_i = \{ W_i^{E}$, $W_i^{D}\}$ as the set of parameters for each layer, and $\Theta=\{\theta_0, \cdots,\theta_l\}$ as the total number of parameters added to the model. 

\paragraph{Plug-and-Play Adapters}
At decoding time, PPLM requires a fixed number of iterations $p$ to generate a single token. This makes the model impracticable for interactive tasks such as conversational models. To cope with this issue, we propose to first use PPLM to generate datasets of dialogues with certain attributes $a$, denoted as $\mathscr{D}^{a} = \{\mathcal{D}^1, \dots,\mathcal{D}^n\}$, and then to optimize the residual adapter parameters to steer the output of the original \texttt{LM} distribution. Hence, for each attribute $a$, we optimize the parameters in $\Theta_a$ to minimize the negative log-likelihood over the dataset of dialogues $\mathscr{D}^{a}$. Formally,
\begin{equation}
   \mathcal{L}(\mathscr{D}^{a}) =  - \sum_k^{|\mathscr{D}^{a}|} \sum_{i}^{n} \log p(s^k_i|s^k_{<i},\mathcal{D}^k_t),
\end{equation}
where each response $S^k_t=\{s^k_0,\cdots,s^k_n\}$ is of maximum length $n$.

\section{Experimental Setup}
In this section, we conduct extensive experiments on the proposed methodology using both automatic and human-evaluation. Differently from PPLM~\cite{dathathri2019plug}, where a set of pre-defined prefixes are used to trigger the generation, in our experiments we use 100 conversations~\cite{adiwardana2020towards} for generating 1100 possible prefixes (i.e., moving window of size two). These open-domain generic dialogues serve as a prefix to trigger the responses rather than fine-tuning. 
In all our experiments, we use DialoGPT~\cite{zhang2019dialogpt} medium, a large pre-trained model trained on 147 Million multi-turn dialogues from Reddit, spanning from 2005 to 2017. Importantly, the proposed methodology is model agnostic, and thus it can be applied to any other large pre-trained model such as Meena~\cite{adiwardana2020towards} and Blender-Bot~\cite{roller2020recipes}. 
Since Plug-and-Play Adapters use the generated responses from PPLM, we randomly split the prefixes with 80\% for learning the adapter perturbation and the remaining 20\% for the final automatic and human evaluation. 
This is done to have a fair comparison between other baselines and adapters (See Appedix A for more details). 

\subsection{Attribute Models}
We train three discriminators covering six attribute models such as Positive, Negative, Question, Sci/Tech, Business and Sport. For controlling positive and negative responses, we use SST-5~\cite{socher2013recursive} with the class Very-Positive and Very-Negative as the attribute. For controlling for Question, we use the speech-act annotation from Daily Dialogue~\cite{li2017dailydialog} with the Question class as the attribute. To avoid any dialogue related data, we only use the sentences without the corresponding context. Finally, for generating the response about Sci/Tech, Business and Sport, we use the AG-NEWS~\cite{zhang2015character} topic-classification dataset, using the respective classes as attributes. As mentioned in Section~\ref{subsec:PPLM}, we freeze the DialoGPT parameters and we train a linear classifier on top of the representations from the final layer of its Transformer blocks. Table~\ref{tab:discriminator}, shows the sample size statistics and the performance in terms of F1-score for all the aforementioned datasets. We also report the current state-of-the-art, to show that a linear classifier trained on top of the DialoGPT activation can reach competitive performance. 

\subsection{Baselines}
We compare multiple plug-and-play settings such as: \textbf{DG}: DialoGPT proposed by~\citet{zhang2019dialogpt}; \textbf{WD}: DialoGPT plus a word level weight-decoding schema as in~\cite{ghazvininejad2017hafez,see2019makes}; \textbf{PP}: DialoGPT plus PPLM~\cite{dathathri2019plug}, as explained in Section~\ref{subsec:PPLM}; \textbf{AD}: DialoGPT with one adapter per style, as explained in Section~\ref{subsec:Adapters}. In all the baselines, we sample 10 different hypotheses using multinomial-sampling after a top-k filtering (with $k=10$), to ensure response diversity~\cite{zhang2020trading}, and we select the hypotheses with the lowest attribute model loss as the response. This re-ranking technique has shown to be very effective for generating good responses~\cite{adiwardana2020towards,dathathri2019plug}. 

\begin{figure*}[t]
    \centering
    \begin{subfigure}[b]{0.45\textwidth}
         \centering
         \includegraphics[width=\textwidth]{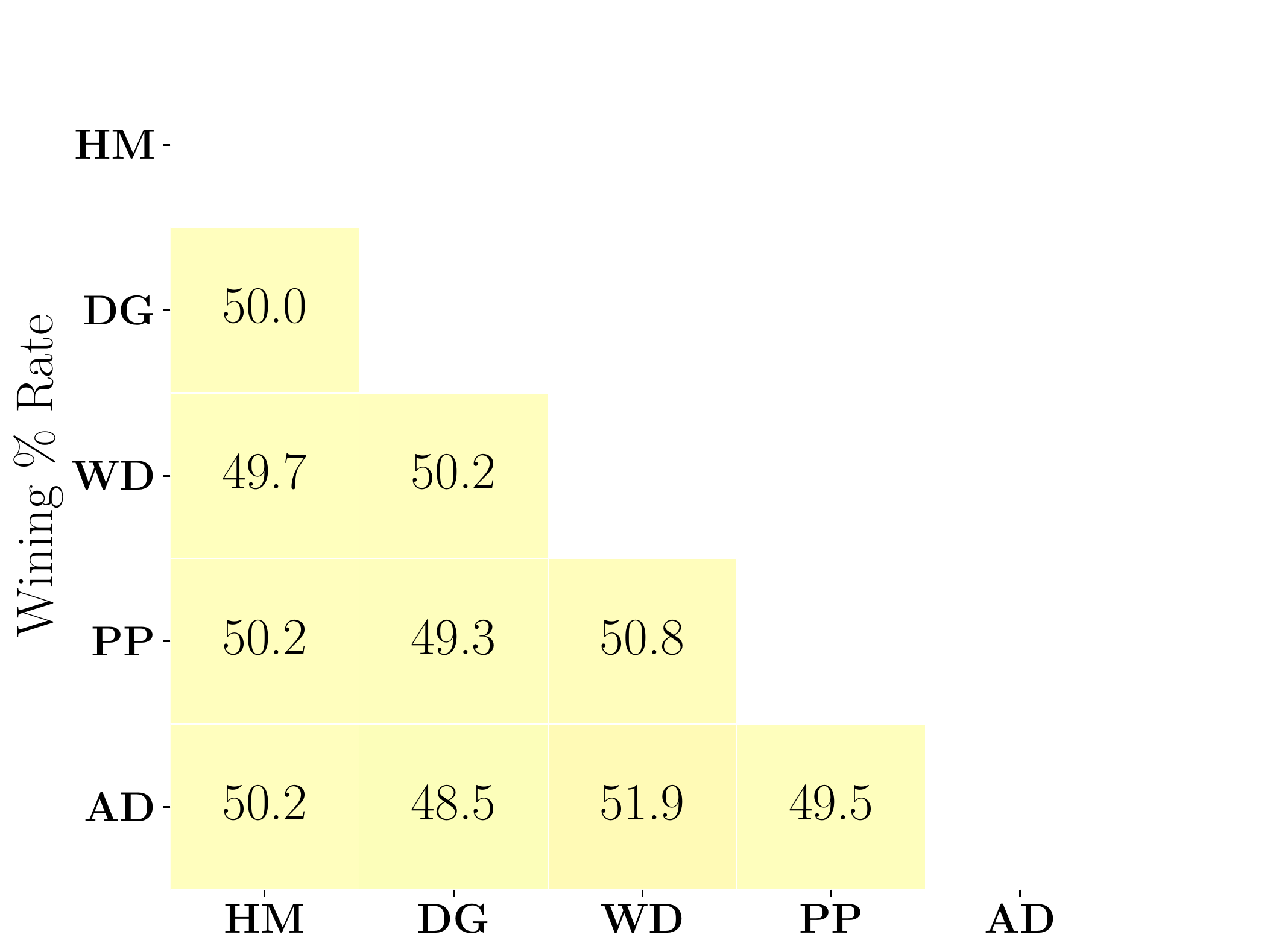}
         \caption{Humanness}
         \label{fig:very_negative}
     \end{subfigure}
    %  \hspace{0.03\textwidth}
     \begin{subfigure}[b]{0.45\textwidth}
         \centering
         \includegraphics[width=\textwidth]{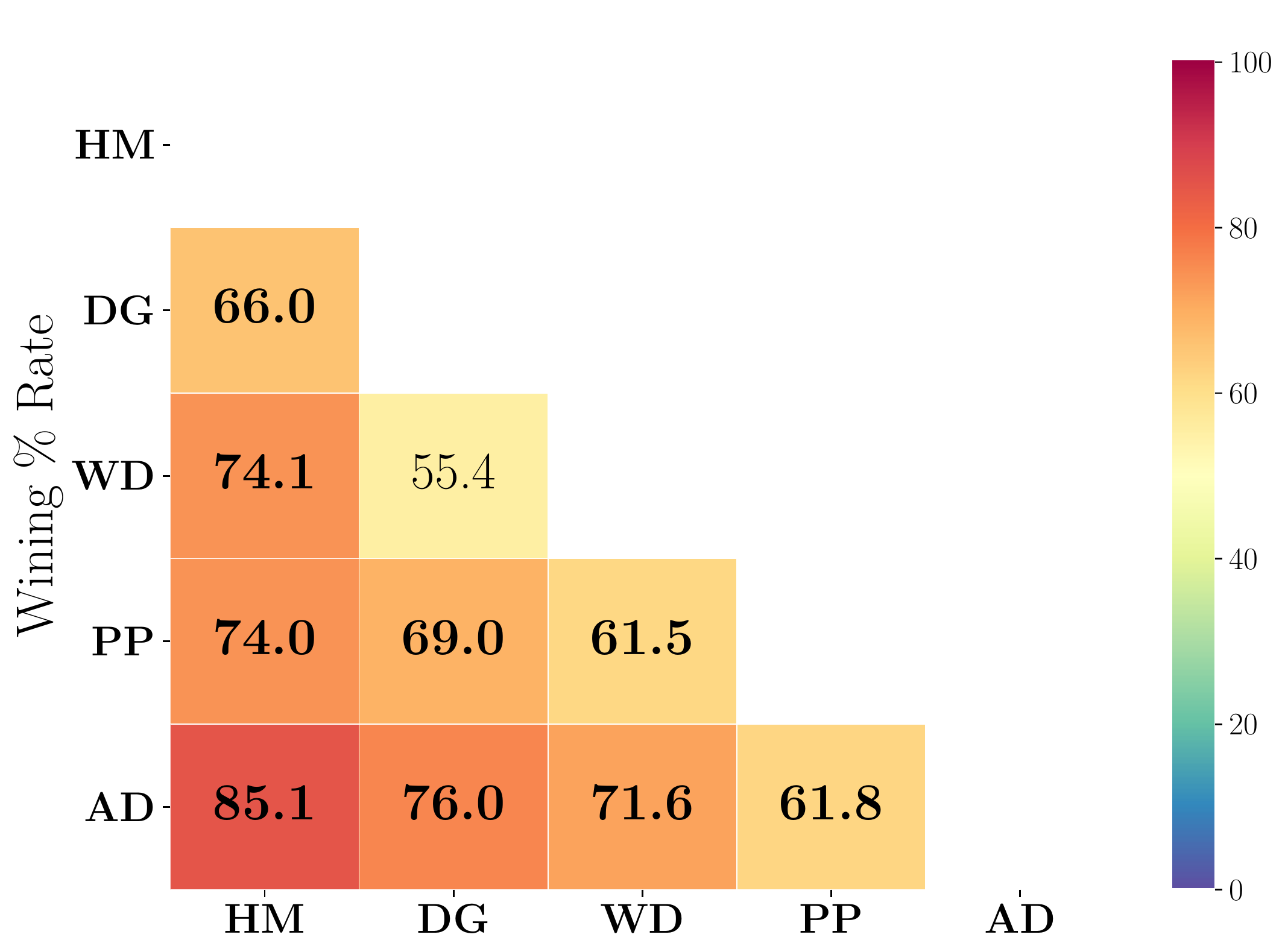}
         \caption{Attribute Consistency}
         \label{fig:very_positive}
     \end{subfigure}
    \caption{Human evaluation results in term of winning rate for both Humanness and Attribute Consistency. For example, in the Attribute Consistency table, \textbf{DG} wins 66\% of the time versus \textbf{HM}. Bold results are statistically significant ($p<0.05$).}
    \label{fig:human}
\end{figure*}

\subsection{Evaluation Metrics}
We evaluate the generated responses using both automatic and human evaluations. 

\textbf{Automatic Eval.} in open-domain chat is challenging~\cite{liu2016not}, especially when using n-grams methods over single reference (e.g., BLEU~\cite{papineni-etal-2002-bleu}). In this paper, no gold-reference response is provided (e.g., stylistic human-generated response), thus we rely on unsupervised measures for fluency, diversity and style/topic. For fluency, we compute the perplexity score of the dialogue prefix plus the generate response using GPT2~\cite{radford2019language}. For diversity, we use the distinct n-grams~\cite{li2016diversity} (normalized by the length of the text) across all the responses generated by a given method. For evaluating the attribute consistency, we train external classifiers using no-overlapping data with the attribute model. For sentiments, we use AMAZON-5~\cite{mcauley2013hidden} product reviews. For topics, we use the test-set data of AG-NEWS~\cite{zhang2015character} because we could not find another topic classification dataset with the same classes. For each dataset, we trained a separate BERT~\cite{devlin2019bert} (base) classifier with a simple classification head. Table 2 in Appendix B, summarizes the dataset statistics and the performance of the trained scorer.

\textbf{Human Eval.} is the most effective way for evaluating open-domain chat-bots. In this paper, we evaluate two aspects from the generated response: Humanness and Attribute Consistency. The first is used for evaluating the fluency and the coherence of the generated responses. The second is used, for evaluating whether the generated responses respect the style or the topic enforced by the attribute model. We use Acute-Eval~\cite{li2019acute} style A/B testing, in which we compare all possible models' pairs (e.g., PP vs. DG etc.). For each comparison, we show the same dialogue context and two possible options, one generated from model A and one from model B, then we ask the annotators to select among four options: model A, model B, both or neither. We collect annotations for both Humanness and Attribute Consistency on 30 dialogues per model comparison and attribute, which amount to a total of 4200 human annotations. Further details are provided in Appendix C.

\begin{figure*}[t]
    \centering
    \begin{subfigure}[b]{0.45\textwidth}
         \centering
         \includegraphics[width=\textwidth]{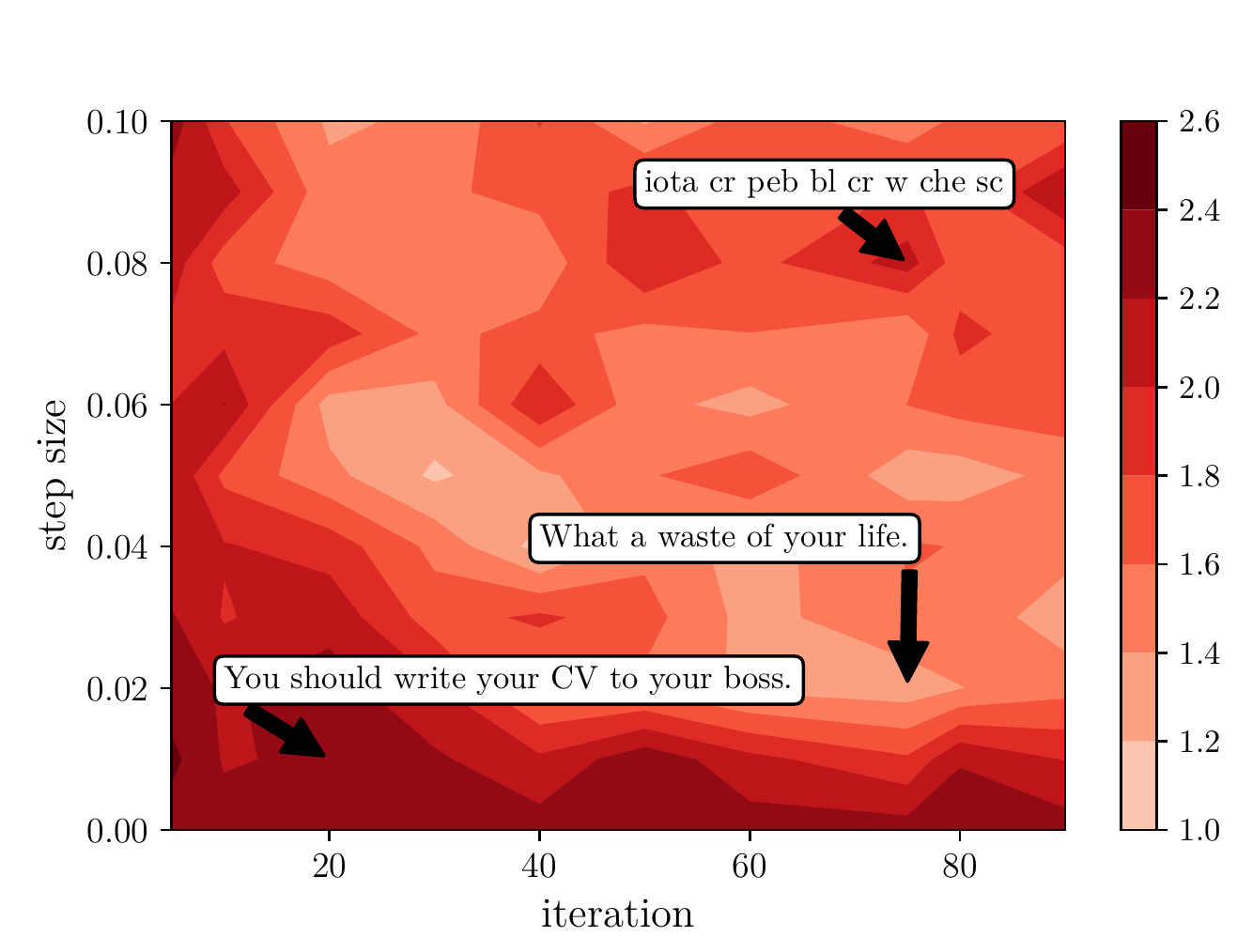}
         \caption{Negative}
         \label{fig:very_negative_iter}
     \end{subfigure}
     \hspace{0.03\textwidth}
     \begin{subfigure}[b]{0.45\textwidth}
         \centering
         \includegraphics[width=\textwidth]{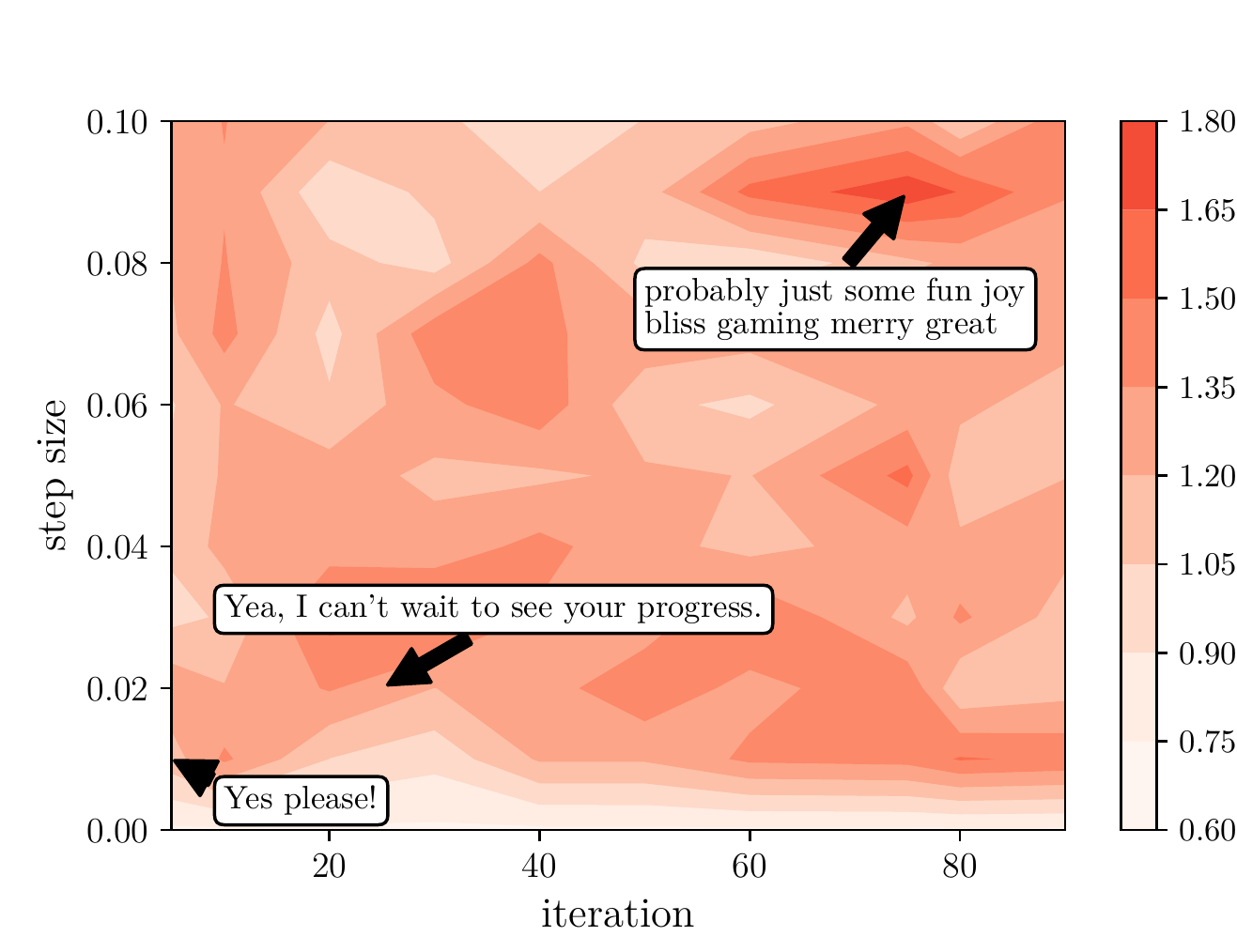}
         \caption{Positive}
         \label{fig:very_positive_iter}
     \end{subfigure}
    \caption{Contour plot of the normalized sum of the log Perplexity score, computed by GPT2~\cite{radford2019language} and the external classifier loss on the generated response by PPLM for the negative and positive style. On the $x$-axis the number of iteration $p$ and on the $y$-axis the step size $\alpha$. Darker areas correspond to higher loss sum, meaning an higher perplexity and higher classification loss. The label represent a sample response from a given iteration and step size. }
    \label{fig:contourplot}
\end{figure*}
\section{Results}
In this section, we evaluate the proposed methodology to answer three research questions:  \textbf{1)} is it possible to use plug-and-play methods for controlling the output of a large pre-trained conversational model? if so, \textbf{2)} what are the most effective plug-and-play methods?, and \textbf{3)} how difficult is to control the response generation given various attributes? To answer the first two questions, we rely on both automatic and human evaluation. Table~\ref{Tab:auto} and Figure~\ref{fig:human} reports the aggregated result for all the styles and topics in both evaluations. The breakdown per attribute is reported in Appendix D. 

\subsection{Quantitative Evaluation}
\label{sec:quant_eval}
\textbf{Automatic Eval.} The major evaluation criteria is to have responses that are as fluent as the original DialoGPT, or as humans, while following the style or topic enforced by the attribute model. In Table~\ref{Tab:auto}, we can see that DialoGPT (DG) achieves the lowest perplexity, but it also has the lowest aggregate attribute score (i.e. Score in the Table~\ref{Tab:auto}). By analysing the breakdown by style, we can see that by default, the original model has a higher score in both positive style and Sci/Tech topic. 
%\Sumanth{should comma be here? I changed the next part. Can someone check? I am sleep deprived}
We hypothesize that this this is due to two factors: 1) The discussions in Reddit are more often related to Sci/Tech topics. By providing general questions as input, e.g., ``What do you do for living?", the model often generate tech related responses, e.g., ``I am a computer science student". 2) The authors of DialoGPT~\cite{zhang2019dialogpt} filtered undesired and toxic responses from the Reddit conversations used in training, which explains the positivity of the DialoGPT responses.
% For instance, they removed all offensive responses using string matching over a large dictionary, and they removed sub-Reddits containing offensive language \Sumanth{Remove this sentence, or merge with last one}.

Using weight decoding (WD) on top of DialoGPT leads to an improvement in both the diversity score and the external classifier score. However, WD tends to increases the perplexity score, showing that the generation fluency with respect to the context is lost. In preliminary experiments, we notice that weight decoding generates responses that are not related to the dialogue context but are highly similar to the distribution of the discriminator datasets. 
This is consistent with the observations in \cite{see2019makes} that 
weighted decoding is difficult to tune and often 
provides control at the cost of fluency, 
leading to non-sensical generation.
On the other hand, PPLM (PP) is able to achieve a lower perplexity compared to WD while attaining both, a higher attribute consistency score and a high response diversity (dist). 
We hypothesize that this improvement is due the ability of PPLM to dynamically perturb the latent activation of the model without breaking the original distribution thanks to the KL regularization and to the Post-norm Geometric Fusion~\cite{stahlberg2018simple}. 

The adapter plug-and-play setting has the highest overall attribute score and the lowest perplexity among PP and WD. However, the response diversity, especially dist-1, is lower than for other baselines, meaning that the response may contain repetitive tokens (e.g., ``so so bad"). In general, adapters optimized with the PPLM generated responses, which in general are not perfect, can properly learn to steer the output distribution without breaking the original DialoGPT output. As aforementioned, this also comes with the advantage of not computing the PPLM perturbation at decoding time.   

\begin{table*}[t]
    \centering
    \resizebox{\textwidth}{!}{
        % \begin{tabular}{llll}
        % \hline
        % \multicolumn{1}{l}{} &
        %   HUMAN 1 &
        %   \multicolumn{2}{l}{Are you doing any home decorating then?} \\
        % \multicolumn{1}{l}{} &
        %   HUMAN 2 &
        %   \multicolumn{2}{l}{Yes! We set up an eco-friendly (i.e. fake) Christmas tree and put up some colorful LED lights which is very festive.} \\ \hline
        %  &
        %   \multicolumn{2}{c|}{\textbf{NEGATIVE}} &
        %   \multicolumn{1}{c}{\textbf{POSITIVE}} \\ \hline
        % \multicolumn{1}{l|}{\textit{DG}} &
        %   \multicolumn{2}{l|}{\begin{tabular}[c]{@{}l@{}}I had a pretty solid day. Just watched the movie, \\ ate pizza.\end{tabular}} &
        %   I'm pretty well. Thanks for asking though! Haha \\ \hline
        % \multicolumn{1}{l|}{\textit{WD}} &
        %   \multicolumn{2}{l|}{\begin{tabular}[c]{@{}l@{}}I'm good too. My job doesn t allow me to be at \\ home today.\end{tabular}} &
        %   \begin{tabular}[c]{@{}l@{}}It feels nice to know people like you exist. I hope\\  things turn well soon!\end{tabular} \\ \hline
        % \multicolumn{1}{l|}{\textit{PP}} &
        %   \multicolumn{2}{l|}{\begin{tabular}[c]{@{}l@{}}I'm pretty bad : I haven't had any work done yet so I guess \\ that's why I don't feel like I've accomplished anything.\end{tabular}} &
        %   That was beautiful! \\ \hline
        % \multicolumn{1}{l|}{\textit{AD}} &
        %   \multicolumn{2}{l|}{\begin{tabular}[c]{@{}l@{}}I had a really awful horrible day yesterday. \\ It's been awful all week. So, so, so, so horrible.\end{tabular}} &
        %   \begin{tabular}[c]{@{}l@{}}That was beautiful! I am very happy for your birthday! \\ Hope you enjoyed it! :D\end{tabular} \\ \hline
        % \end{tabular}
        \begin{tabular}{llll}
    \hline
     &
      HUMAN 1 &
      \multicolumn{2}{l}{Are you doing any home decorating then?} \\
     &
      HUMAN 2 &
      \multicolumn{2}{l}{Yes! We set up an eco-friendly (i.e. fake) Christmas tree and put up some colorful LED lights which is very festive.} \\ \hline
     &
      \multicolumn{2}{l|}{\textbf{NEGATIVE}} &
      \textbf{POSITIVE} \\ \hline
    \multicolumn{1}{l|}{\textit{DG}} &
      \multicolumn{2}{l|}{\begin{tabular}[c]{@{}l@{}}That sounds awesome. I'm going to get a couple of these \\ and start decorating.\end{tabular}} &
      Oh that's so cool! I love your stuff!! \\ \hline
    \multicolumn{1}{l|}{\textit{WD}} &
      \multicolumn{2}{l|}{\begin{tabular}[c]{@{}l@{}}I'm not sure if you're aware of this, but the word \\ ecoreneive actually refers to being ecoreneous...\end{tabular}} &
      Very nice, good sir \\ \hline
    \multicolumn{1}{l|}{\textit{PP}} &
      \multicolumn{2}{l|}{I'm not a fan of LEDs in general. They always seem to fail.} &
      Oh wow awesome! Thank you so much for your time! \\ \hline
    \multicolumn{1}{l|}{\textit{AD}} &
      \multicolumn{2}{l|}{That sounds like the absolute most boring thing. EVER.} &
      That is amazing! I am so excited!! :D So creative and creative!! :D \\ \hline
    \end{tabular}
    }
    \caption{Examples of generated responses for negative and positive with the same starter. }
    % PP, AD uses rest\textit{DG} tends to be positive, which leads easier controlled generation for positive than negative.}
    \label{tab:responses_neg_pos}
\end{table*}
\textbf{Human Eval.} In Figure~\ref{fig:human}, we report the winning rate of the A/B testing for both humanness and attribute consistency. From these tables, we can highlight: 1) There is not statistically significant difference in the humanness score among the multiple methods, even with 210 annotations per cell. In general, all the methods lose with the human response (HM), but not by a large margin. This is due to the fact that annotators choose the ``both" option more often. 2) In term of attribute consistency, we observe that the methods form a clean, well-ordered rank such as \textbf{AD}$>$\textbf{PP}$>$\textbf{WD}$>$\textbf{DG}$>$\textbf{HM}, which confirms the automatic evaluation results. Different from humanness, all the results except WD vs. DG are statistically significant ($p<0.05$), showing the adapter clearly defeats other methods. 

To answer the first two research questions, we observe that both automatic and human evaluation show that plug-and-play methods are suitable for controling response generation. Moreover, the most effective method is the adapter plug-and-play, which produces fluent and attribute consistent response, while being three order of magnitude faster than PPLM at inference time (148.5s/token vs. 0.123s/token) using a single Nvidia 1080Ti.

\section{Analysis}
\label{sec:analysis}
In this section, we evaluate the difficulty of controlling the response generation for a given attribute.
To do so, we analyse the behaviour of PPLM over two opposite styles (i.e., positive and negative) and then we conduct a qualitative evaluation over the generated responses.

\paragraph{Iteration \& Step Size}
We analyse the loss of the automatic scorer for fluency and attribute consistency to understand the effects of the number of iterations $p$ and the step size $\alpha$ in Equation~\ref{pplm}. Figure~\ref{fig:contourplot} depicts the normalized sum of the log Perplexity score, computed by GPT2~\cite{radford2019language} and the external classifier loss on the generated response for the negative and positive style. In general, the aggregate loss for the negative attribute (Figure~\ref{fig:very_negative_iter}) is higher than the positive attribute (Figure~\ref{fig:very_positive_iter}), as also shown in the sampled responses, where small steps size and few iterations leads to positive responses. However, when both the step size and the iteration surpass a certain threshold, the conditioning becomes very strong and the text generated by PPLM loses its fluency. Overall, this visualization suggests that it is more laborious to control for the negative sentiment with PPLM, and there is a smaller region for the hyper-parameters space where the responses are both fluent and attribute consistent.

% suggesting that it is more laborious to enforce PPLM to be negative. In fact, as shown in the sampled responses in Figure~\ref{fig:very_negative}, little step and iteration lead to a low loss for the positive discriminator, while  response yet not following the stylized response. 

\paragraph{Qualitative Analysis}
We sample and read 200 dialogues responses from the adapter plug-and-play model (AD), and we study the overall quality of the response especially to understand when and why DialoGPT is hard to steer. We discover three possible factors:
%\textbf{1)} the context influences the hardness of the response steering, 
\textbf{1)} the context influences the hardness of the response steering, %additional restrictions from the given dialogue context,
\textbf{2)} available vocabulary for attributed style/topic, and \textbf{3)} mutual exclusivity of the attribute-specific vocabulary.

% % If the distributions $p(X|a)$, adapter plug-and-play, and $p(S)$, the original is closer, it requires less steering to get an attribute optimized response.
% Besides attributes, the dialogue prefix restricts the possible response of DialoGPT, thus make in it harder to steer compare to language models as in~\citet{dathathri2019plug}. that use short prefixes (e.g., ``The issues ..") to trigger the generation~\citet{dathathri2019plug}
\textbf{1)} Unlike language models that use short prefixes (e.g., ``The issues ...") to trigger the generation~\citet{dathathri2019plug}, conversational models are constrained to the given dialogue history which significantly influences the controllability. 
% Besides attributes, the given dialogue prefix significantly influences the controllability. Since the dialogue prefix restricts the possible generation directions more strictly, DialoGPT is harder to steer than language models in~\citet{dathathri2019plug}.
%The given dialogue prefix influences more significantly the controllability of the conversational model than a language model in~\citet{dathathri2019plug}, since the prefix restricts the possible generation directions more strictly. 
% Given an attribute-related context (e.g., Table 13 in Appendix), AD generates an impressively natural and on-topic response, but when provided a less natural context for the attribute (e.g., Table 17 in Appendix), AD generates a response that may sound sudden and out of context.
% As also discussed in Section~\ref{sec:quant_eval} and can be seen in Table~\ref{tab:responses_neg_pos}, DialoGPT tends to generate a positive and/or techy response by default, removing obstacles in forcing these attributes.
Given an open ended dialogue context (e.g., Table 11 in Appendix), AD generates an impressively natural and on-topic response, but when provided a more constrained dialogue context (e.g., Table 17 in Appendix), AD generates a response that may sound sudden and out of context.

\textbf{2)} Looking at the overall responses, also shown in Table~\ref{tab:responses_neg_pos}, we observe that models use a restricted vocabulary for generating attribute consistent responses. For example, AD frequently generates sentences containing ``horrible", ``terrible" or ``worst" for negative, while ``beautiful", ``happy" or ``wonderful" are more common for positive. 
% In Appendix Table 3 and 4, we report an attribute-specific vocabulary. 

\textbf{3)} The importance of mutual exclusivity of the attribute-specific vocabulary also explains the relatively poor performance when controlling for certain topics. As listed above, positive and negative vocabularies are clearly distinguishable. However, the attribute-specific words for topics such as Business are more generic (e.g., ``car", ``store") than other topics such as Sport (e.g., ``football", ``hockey") or Sci/Tech (e.g., ``android", ``software"). If the attribute-specific words are common and shared across multiple domains, the generated responses may not sound attribute specific even though the correct vocabulary is used. 

Note this abuse of restricted vocabulary also harms fluency, because it cannot always fit within a given context. Additional generated examples and statistics of attribute-specific vocabulary on each style/topic are provided in Appendix D. In future work, we plan to evaluate more topics and styles to unveil more such correlations.

% It implies that the sentiment attributes are easier than topic attributes, since there are only three (positive/negative/neutral) categories thus their vocabulary is larger.

\section{Conclusion}
We explore plug-and-play methods for controlling the response generation of large pre-trained conversational models in a light-weight manner while being effective. With extensive automatic and human evaluations, we show that PPLM is able to generate fluent and attribute consistent responses. Further, to overcome the significant computational overhead introduced by PPLM at decoding, we optimize a tiny residual adapter for each attribute based on a few synthetic responses generated using PPLM. The resulting model does not require further computation at decoding time, and outperforms PPLM both in terms of fluency and attribute consistency.

\section{Acknowledgements}

This work has been partially funded by MRP/055/18 of the Innovation Technology Commission, The Hong Kong SAR Government. We would like to thanks Jason Yosinski
and the MLCollective for the helpful discussion.

\bibliography{anthology,emnlp2020}
\bibliographystyle{acl_natbib}

\end{document}